# An Optimal Experimental Design Approach for Light Configurations in Photometric Stereo

*Hamza Gardi* ✉ [1], *Sebastian F. Walter* [2] *and Christoph S. Garbe* [1]

Abstract

This paper presents a technique for finding the surface normal of an object from a set of images obtained under different lighting positions. The method presented is based on the principles of Photometric Stereo (PS) combined with Optimum Experimental Design (OED) and Parameter Estimation (PE). Unclear by the approach of photometric stereo, and many models based thereon, is how to position the light sources. So far, this is done by using heuristic approaches this leads to suboptimal and non-data driven positioning of the light sources. But what if the optimal positions of the light sources are calculated for photometric stereo?

To this end, in this contribution, the effect of positioning the light sources on the quality of the normal vector for PS is evaluated. Furthermore, a new approach in this direction is derived and formulated. For the calculation of the surface normal of a Lambertian surface, the approach based on calibrated photometric stereo; for the estimation the optimal position of the light sources the approach is premised on parameter estimation and optimum experimental design. The approach is tested using synthetic and real-data. Based on results it can be seen that the surface normal estimated with the new method is more detailed than with conventional methods.

*Keywords*: Surface normal Estimation. Photometric Stereo. Optimum Experimental Design. Parameter Estimation. Lambertian Surface. Calibrated Photometric Stereo.

## 1. Introduction

Surface normal is one of the major ways to parameterize an extensive type of surface. They are required for the visualization of computer graphics and used in the field of visual inspection in the industrial environment. There are two methods for determining the normal vector: geo-metric methods and photometric methods. The normal vector is defined as a derivative of the surface geometry, i.e. an error in the geometry of the surface has an effect on the normal vector. In order to decouple the determination of the normal vector from the geometry of the sur-face, photometric methods for determination of the nor-mal vector is used. One of the most important technique of photometric methods in practice is photometric stereo.

Photometric Stereo (PS) aims to estimate the surface normal and reflectance at every point of an object by using the intensity recorded from multiple images. These images are obtained from the same point of view but under different positions of the light sources. The images of a 3D-object depend on the object´s shape, reflectance properties and the distribution of the light sources; they also depend on the positions of the object relative to the imaging system and on the object´s pose in the space [1]. The brightness of an object is proportional to the irradiance of the illumination source and the value of the Bidirectional Reflectance Distribution Function (BRDF) [2]. The reflectance map is useful as it makes the relationship between surface orientation and brightness explicit; it depends on the BRDF, the surface normal and the distribution of the light source.

In most scientific disciplines, mathematical models for describing systems or processes are fundamental. These mathematical models are often dynamic in nature and often can be described by derived models that are of a simple mathematical form. This process often involves statistical approximation and can be described by differential equation models [3]. In image processing, partial differential equations (PDE) is a widely used method e.g., for image segmentation, image restoration and motion estimation. An overview of the use of PDE in image processing is given in [4]. Usually, the tasks depend on several variables; it is difficult therefore to formulate Optimum Experimental Design (OED) using PDE. Once the task depends on only one variable, it is easier to formulate OED.

To obtain realistic results of mathematical models, the model has to be validated based on experimental data in a process termed model calibration. Model calibration is a method by which a parameter of a mathematical model is estimated using experimentally measured data to obtain a model, by which the best possible real process can be approximated [5]. To fit the model to the data, unknown values in the model, called Parameters $\in \mathbb{R}^n$, have to be estimated by minimizing a norm of the residuals between data and model [3]. Random errors appear in the measurement process. Hence, finding the true parameters is also influenced by information loss due to noise. Only if the statistical uncertainty of the parameter is small, the behavior of the real process can be described in a correct way. The statistical uncertainty of a parameter estimate depends upon layout, setup, control and sampling of experiments. Experimental design optimization problems minimize a function of the Variance-Covariance matrix of the parameter estimation problem [3].

✉ hamza.gardi@gmail.com
[1] Image Processing and Modeling, IWR, University of Heidelberg, Germany
[2] Simulation and Optimization, IWR, University of Heidelberg Germany

The main feature of this article is to compute the optimal positions of the light sources for the calibrated Photometric Stereo of a Lambertian surface. Both uncalibrated PS methods and non-Lambertian property of the objects are beyond the scope of this article. We refer the readers to [6], [7] and [8] for a comprehensive review of uncalibrated, non-Lambertian PS methods. The approaches in this article based on PE and OED, we are developing a new method for calculating the optimal positions of light sources; we use these positions to estimate surface normal by means of a calibrated PS. This article is structured as follows: After this introduction, part 2 gives an overview of related works. In part 3, section 1, the background and theory of PS will be described, whilst in section 2 to 5, PE problems and OED are formulated for our concept. Part 4 deals with the evaluation of our model by means of synthetic and real-data set while part 5 gives the conclusion of this work.

## 2. Related Work

There are many studies, which use photometric stereo (PS) to estimate surface normal and many studies which have applied model-based parameter estimation (PE) and optimum experimental design (OED) by describing processes in science and engineering. However, to the best of our knowledge, this is the first study to combine PS on the one hand, with PE and OED on the other. For this reason, we give a brief overview of each topic separately. Photometric stereo [9] estimates both surface normal and albedo of a Lambertian surface as seen from each pixel in a fixed view, using a set of images taken under different light configurations. These configurations can be time multiplexed, by taking subsequent images for each of the light positions, or frequency multiplexed using differently colored lights. Numerous extensions of the basic schemes of PS have been proposed in literature.

Johnson and Adelson [10] describe an optimization scheme that estimates the surface normal of a diffuse object with constant albedo from a single image under natural but known illumination. Hayakawa describes in [11] a method to estimates both surface normal and surface reflectance of Lambertian objects. His method works without prior knowledge of light source direction. He uses Singular Value Decomposition (SVD) to factorize the image data matrix, which is the product of surface matrix and light source matrix.

Zickler et al [12] use Helmholtz stereopsis by acquiring images in which the positions of the light source and camera are replaced. This enables the recovery of surface normal and depth; however, this method requires finding the corresponding points in images, which are taken from different directions of view.

Wang and Dana [13] and Ma et al [14] conceived a method to compute surface orientation for shying objects under the assumption that surface normal and halfway vector are adapting, when the maximum value of reflectance is obtained. In [15] Wang et al analyzes for reconstruction error in the near lighting PS by optimizing the baseline, the distance between two light sources in the ring-light setup for PS. The concept of near lighting PS is also discussed in [16], the authors use PS technique combined with stereo camera to determine the depth map of an object. The authors in [17] propose a new approach based on convolutional neural network architecture for general BRDF photometric stereo. For given images and corresponding light directions, surface normal and BRDF of the surface can be computed. The optimized weights of the network are achieved by minimizing the reconstruction loss between testing and observed images.

Nielsen et al [18] introduced a method to reconstruct a BRDF of objects by computing the optimal directions for both the light sources and the cameras. The method performs minimizing the condition number of the principal component matrix of the data.

Ray et al performs in [19] error analysis of normal vector determined by radiometric methods. They identify the error sources and proposed solutions and strategies to deal with them. Direction of the light sources is one of the important error sources. In order to minimize the error by radiometric methods, the best configuration of light sources has to be detected. Argyriou et al [20] use $l_1$-norm to estimate the optimal directions of the light sources, by minimizing the shadow effects. Their method requires each class of objects a training data set and four light sources. Another work of Argyriou [21] presents two approaches to determine the optimal directions of the light sources. The first method requires the distribution of the surface normals and four light sources to estimate optimal directions of the light sources by minimizing the effects of shadows and highlights and maximizing the surface details at the same time. The second method is based on the density of the iso-contours in the gradient space.

Compared to [20] and [21], we use the principles of PE and ODE to calculate the optimal configurations of the light sources; we use only 3 light sources.

In [22] Drbohlav and Chantler present a method to find out the optimal configurations of the light source in PS. The method is based on reducing the camera noise and it is independent of the shape of the illuminated object.

Spence and Chantler introduces in [23] a method based on the sensitivity analysis of PS to reduce the noise ratio in each image to determine the surface normal. Again, the method presented is independent of the shape of the illuminated object. In [24], the authors extend the method in [23] to textured objects.

Our work differs from [22], [23] and [24] by four significant aspects. First, for the calculation of the optimum position of the light sources our method performs an optimization that depends on the shape of the illuminated object. Second, the light sources of our method are distributed independently of each other in 3D space. The third and most important one is that our approach is based on PE and ODE. Fourth, we minimize the confidence region around normal vector to determine the optimal configurations of the light source. To compare the accuracy of our method with the method presented in [22], we describe, in section 4.1, the error analyses of the surface normals of both methods.

In [3] an overview of PE and ODE is given based on a model. The authors use generalized Gauss-Newton for PE; they also minimized the statistical uncertainty of the parameter to fit models to data and to validate the parameters.

In [25] an efficient optimization method for experimental design is presented. The method uses a combination of discrete adjoint computations, Taylor arithmetic and matrix calculus, thereby achieving a faster processing time and a reduction of complexity. An application of PE and OED in industry is presented in [26]. They used derivative based parameter identification to improve the precision by calculation of motion (e.g., movement of robot arm) to find the optimal measurement configuration and for the parameter identification they minimize the variance-covariance matrix of measurement configurations. This reduces probability of collision. An application of the developed approach was tested with 'Tool Center Point' (TCP).

The authors of [27] use the concept of OED to formulate an active learning method to select the most informative samples in the database for image retrieval; they called it Geometric Optimum Experimental Design (GOED). Compared to the popular active Support Vector Machine (SVM) their method is label independent.

Chapter 15 of [3] gives a brief overview of the application of PE in some areas of image processing and computer vision. The methods of PE can be used to estimate the entries of filter kernel in low-level image processing, such filtering is used for computing first- and second- or higher derivative filters.

In the field of optimal control in image processing, Chen and Lorenz in [28] use optimal control to select useful regions in the images for image sequence interpolation. The principle of optimal control is also used in [29] to compute the transformation of an image for image registration and in [30] to determine optical flow precisely.

## 3. Background and Theory
### 3.1. Photometric Stereo

The combination of scene illumination, surface reflectance and surface orientation in view centered coordinates is called the Reflectance Map [1]. It specifies the brightness of a surface at a particular orientation for a given distribution of illumination and surface material. The reflectance map is independent on the viewed shape of the object and it represents knowledge about image intensity. Horn [1] was the first to formulate the Shape of Shading (SFS) problem which amounts to the solution of a nonlinear Partial Differential Equation (PDE). This relation is called brightness equation.

$$I(u,v) = Ref(N(u,v))$$

(1)

Where $(u, v)$ represents the coordinates of the point on the image plane, $Ref$ the reflectance map and $N$ the surface normal. The brightness equation connects the reflectance map with the brightness of an image. The SFS equation is known to be an ill-posed problem as the solution of this PDE equation is not unique [1] and it needs additional information to solve this equation uniquely. A typical way to solve the SFS equation with one image is to make assumptions such as a priori constraints on the reflectance map, a priori constraints of surface curvature, or global smoothness constraints. In contrast, photometric stereo uses additional images instead of a priori assumptions. Woodham [9] use three-point light sources captured by three pictures to solve the SFS equation uniquely and he therefore developed a new technique termed 'photometric stereo'.

The PS method is one technique for solving SFS. The fundamental theory follows the principles of optics. An image irradiance equation is developed to calculate image irradiance as a function of the surface orientation. This relationship cannot be inverted because image brightness is only one measurement, whereas the surface orientation has two degrees of freedom [9]. In orthographic projection, for all points of any object surface, the viewing direction and hence the phase angle is constant. Therefore, for a fixed light source and viewer, the ratio of scene radiance to irradiation forms an image intensity which depends only on gradient coordinates [9]. With the known values of intensity and illumination direction, PS makes it possible to estimate both local surface orientation and local surface curvature without demanding either global smoothness or prior image segmentation [31]. Assuming an orthographic projection with the camera direction aligned with the negative $Z-axis$, as represented by Woodham in [9], a relationship between intensity and surface orientation is established. For any object point $(x, y, z)$ the brightness equation will map onto an image point $(u, v)$ where $u = x$ and $v = y$.

If the object surface is given as $z = f(x, y)$ then the surface normal is defined in vector form as $[\partial f(x,y)/\partial x, \partial f(x,y)/\partial y, -1]^T$. If parameters $p$ and $q$ are defined by $p = \partial f(x,y)/\partial x$ and $q = \partial f(x,y)/\partial y$ then the surface normal can be written as $[p, q, -1]^T$. $(p, q)$ is a gradient of the function $f(x, y)$. The gradient space is a practical way to represent surface orientation. It has been used in both scene analysis and image analysis; in image analysis it is used to join the geometry of image projection to the radiometry of image formation [9]. Not all incident light is radiated from a surface. This radiometric effect can be incorporated in to the image irradiance equation with an albedo factor:

$$I(u, v) = \rho\, Ref(p, q).$$

(2)

With the albedo factor $\rho$ such that $0 < \rho < 1$. Surface albedo $\rho$ is a number capturing the surface reflection property at location $(x, y)$. As shown in [9] the brightness of a diffuse surface illuminated by a point light sources depends on the $\cos$ine of the angle between the surface normal and the light source. Then the registered image intensity $I(u, v)$ of a point on object surface is given by

$$I(u, v) = I\, \rho \cos\emptyset = \rho\left(N_x S_x + N_y S_y + N_z S_z\right),$$

(3)

where the surface normal $N$ and the light source direction $S$ are given by $N = (N_x, N_y, N_z)^T$ and $S = (S_x, S_y, S_z)^T$. Instead of one image of an object let us take three images of the same object without changing either the camera or the object, but with different positions of light sources turned on at the time. These three different light source directions, relative to the camera are described by vectors $S^1 = (S_x^1, S_y^1, S_z^1)^T$ $S^2 = (S_x^2, S_y^2, S_z^2)^T$ $S^3 = (S_x^3, S_y^3, S_z^3)^T$. Corresponding pixels in the three images would have three different intensities $I_1, I_2, I_3$ and the corresponding light sources directions. Let the normal corresponding to the pixel be denoted by $N = (N_x, N_y, N_z)^T$. Assuming a Lambertian surface, the three intensities can be collated to the surface normal and the light source directions

$$\underbrace{\begin{pmatrix} I_1 \\ I_2 \\ I_3 \end{pmatrix}}_{\substack{\text{in case of}\\ \text{3 light sources}\\ 3X1}} = \rho \underbrace{\begin{pmatrix} S_x^1 & S_y^1 & S_z^1 \\ S_x^2 & S_y^2 & S_z^2 \\ S_x^3 & S_y^3 & S_z^3 \end{pmatrix}}_{\substack{S\\ 3X3}} \underbrace{\begin{pmatrix} N_x \\ N_y \\ N_z \end{pmatrix}}_{\substack{N\\ 3X1}} = m(S, N).$$

(4)

Equation (4) is a forward model, for known surface normal $N$, light source directions $S$ and surface albedo $\rho$, the intensity value can be calculated. But our research is aimed to compute surface normal from image intensity and light source directions. In other words, we have to solve an inverse problem. We describe this in the section after next. Since the process of recording an image containing errors, therefore we specify the Error Model of our model in the next section.

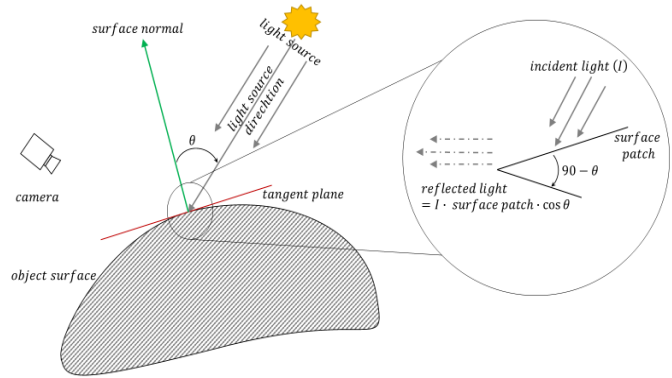

Figure 1 Relationship between the surface and the light source. The Amount of light reflected by a surface patch is proportional to the cosine of the angle between the light source direction and the surface normal.

### 3.2. Error Model

The forward model is a theoretical model; the model would be correct if there is no image noise; by an image acquisition the intensity value is noisy. In this article, we assume the measurement model is defined as follows

$$I_i := m_i\big(S_{true}(t_i), \widetilde{N}_{true}\big) + \epsilon_i \quad , i = 1, \dots, M \;.$$

(5)

In other words, the measured value consists of two parts; first, the model $m_i$ and second, the measurement error $\epsilon_i$. Where $M$ is the number of images taken at times $t_0 < t_1 < \cdots < t_M$. $I_{true}$, $\widetilde{N}_{true}$ and $S_{true}$ stands for true value of intensity, normal vector and positions of the light sources; exact calculation can be carried out with these values.
As discussed in [32] the image noise model obeys Poisson distribution and can be approximated by a Gaussian model as an additive, independent and normally distributed at each pixel [33].

$$\epsilon_i \sim \mathcal{N}(0, \sigma_i^2) \;,$$

(6)

with known variance $\sigma_i^2$, $i = 1, \dots, M$ and covariance $Cov(\epsilon_i, \epsilon_j) = 0 \;;\; i \neq j$.

### 3.3. Inverse Problem

In the Equation (4) the known variables are Intensities $I_1, I_2, I_3$ and the light source directions $S_1, S_2, S_3$. The unknowns are surface albedo $\rho$ and surface normal $N$. If $m = n = 3$, we can recover the surface normal $N$ by normalizing the recovered $\widetilde{N}$ vector, using the fact that the magnitude of the normal is one.

$$\widetilde{N} = S^{-1} I \;,$$

(7)

$$\rho = |\widetilde{N}| \;,$$

(8)

$$N = \frac{\widetilde{N}}{|\widetilde{N}|}.$$

(9)

In case $m \neq n$, we formulate the problem as a linear-least-square based on Parameter Estimation and optimum Experimental Design. In this section we are developing a new method to calculate the best position of light sources. Based on these positions, we calculate the surface normals with PS.

Consider the problem of finding a vector $\widetilde{N} \in \mathbb{R}^n$ such that $S\widetilde{N} = I$, when the matrix $S \in \mathbb{R}^{m \times n}$ and the intensity vector $I \in \mathbb{R}^m$ are given. There are many possible ways of finding the best solution. A choice that can often be motivated for statistical reasons and also leads to simple computational problem. PS is a linear model, one can represent (7) as

$$\underset{\widetilde{N} \in \mathbb{R}^n}{\text{minimize}} \|I - S\widetilde{N}\|_2^2 \qquad S \in \mathbb{R}^{m \times n}, I \in \mathbb{R}^m \;.$$

(10)

Where $S$ has $m \times n$ degrees of freedom, cf. Equation (4). We calculate the optimal positions of the light sources $S$ so that the surface normal $N$ is best possible estimated. In the Equation (2), $S$ and $I$ are mapped to $\widetilde{N}$; $S, I \longrightarrow \widetilde{N}$, the solution of the Equation (2), as shown in [34] is

$$\widetilde{N}' = (S^T S)^{-1} S^T I \;,$$

(11)

where $\widetilde{N}'$ is any minimizing least square solution of the set $I \approx S\widetilde{N}$.

A measurement model representing the relationship between $S, \widetilde{N}$ and $I$ is given by $m(\widetilde{N}, S) = I_{true}$. Both $I$ and $\widetilde{N}$ are random variables. The mapping function

$$I \overset{S}{\underset{\mathcal{G}}{\longmapsto}} \widehat{N} = \mathcal{G}(S, I) \;,$$

(12)

defines, how $\widehat{N}$ derived from measured $I$ and given light sources $S$. Where $\widehat{N}$ is random normal vector. Since $\widehat{N}$ is a random variable, for each measurement we measure a different value of $\widehat{N}$. One can imagine it like a point cloud of $\widehat{N}$ values. Now, the task is to obtain this point cloud as compact as possible. For this purpose, we will use Confidence Region cf. next section.

### 3.4. Statistical Analysis

For the statistical analysis of the solution, the Confidence Region is considered. The Confidence Region $CR$ is a region around a point estimate and indicates the specific probability of having the true value in this region. Now, we declare that the true value of the normal vector is located with a specific probability in the CR region. The variables $I_i$ and $m_i$ are random variables, and the Equation (5) is a Probability Distribution Function (PDF).

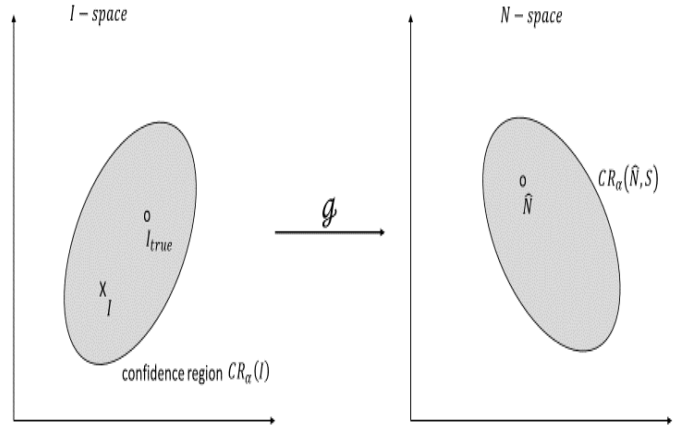

*Figure 2* This figure shows the true value $I_{true}$ is not known, but it is possible to provide confidence region $CR$. The probability that the true value $I_{true}$ is in $CR$ is $1 - \alpha$.

We like to deduce an estimation from the random variables $I_i$ and $m_i$, because $m_i$ involve the normal vector and that´s what we like to compute. We prescribe, that the measurement error $\epsilon_i$ is normally distributed around $(0, \sigma_i^2)$, according to linearity of the multivariate normal distribution [35], follows that $I_i$ is also normally distributed around $(m_i, Id\ \sigma_i^2\ Id^T)$, where $Id$ is identity matrix. Now, we like to derive the distribution of $\widehat{N}$ from the random variable $I$. This is not a simple task as Equation (12) is a non-linear function, so we use Taylor expansion and neglect the high terms.

The Taylor expansion of $\widehat{N} = g(S, I)$ can be written as

$$\widehat{N} = g(S, I) = g(S, I_{true}) + \frac{\partial g}{\partial I}(I_{true})(I - I_{true}) + \underbrace{O(\|I - I_{trure}\|^2)}_{\substack{neglect\ in\ the\ following \\ \left(\substack{assumption\ that\ the \\ problem\ almost\ linear}\right)}}.$$

(13)

The confidence region of $\widehat{N}$ is defined as following

$$CR_\alpha(\widehat{N}; S) = \{N: \mathbb{R}^{n_N}: (N - \widehat{N})\ Cov(S)(N - \widehat{N}) \leq \varkappa_\alpha(n_N)\},$$

(14)

with the covariance matrix $Cov \in \mathbb{R}^{n_N \times n_N}$.

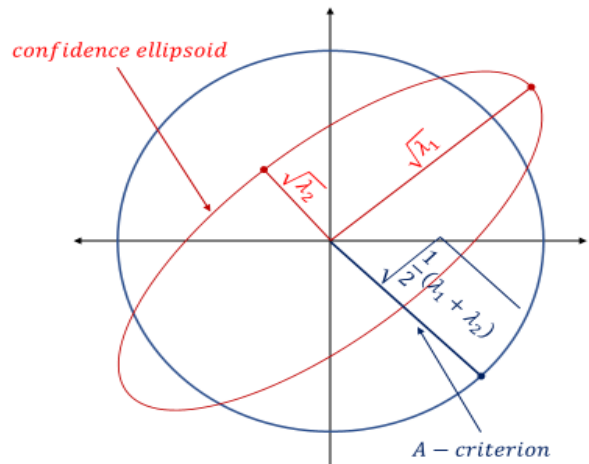

*Figure 3* This illustration shows the relation between OED and confidence region described by the covariance matrix. The eigenvalue of the covariance matrix is $\lambda_1$ and $\lambda_2$.

When we take an image with the camera, the image is always noisy. I.e. each time by solving PE-problem Equation (10), we have another value of $\tilde{N}$. In other words, the noise of $I$ is propagated to the noise of $\tilde{N}$. Naturally we would like to have the value of $\tilde{N}$ as accurately as possible, therefore by means of confidence region, we describe the accuracy of calculating $\tilde{N}$ by the covariance matrix $C$.

$$C = Cov(\tilde{N}) = \mathbb{E}\left[\left(\tilde{N} - N_{true}\right)\left(\tilde{N} - N_{true}\right)^T\right]$$

(15)

$$= \mathbb{E}\left[\left(\mathcal{g}(S, I_{true}) + \frac{\partial \mathcal{g}}{\partial I}(I_{true})(I - I_{true})\right.\right.$$

$$\left.\left.- \mathcal{g}(S, I_{true})\right)\left(\mathcal{g}(S, I_{true}) + \frac{\partial \mathcal{g}}{\partial I}(I_{true})(I - I_{true}) - \mathcal{g}(S, I_{true})\right)^T\right]$$

$$= \underbrace{\frac{\partial \mathcal{g}}{\partial I}(I_{true})}_{K} \underbrace{\mathbb{E}[(I - I_{true})(I - I_{true})^T]}_{\Sigma_I^2} \underbrace{\frac{\partial \mathcal{g}}{\partial I}(I_{true})^T}_{K^T}$$

$$C(S) = K \Sigma_I^2 K^T,$$

(16)

where $K = \frac{\partial \mathcal{g}}{\partial I}(I)|_{I=m(\tilde{N},S)}$ and matrix $\Sigma_I^2 = \text{diag}(\sigma_1, \ldots, \sigma_M)$.

### 3.5. Experimental Design Optimization

The co-variance matrix describes the size of the confidence region. If this region is large, we cannot determine the true value of $\tilde{N}$ precisely. Therefore, the aim is to keep the confidence region as small as possible. For optimization, we need to define what is considered large and what is considered small. It is necessary therefore to define a function that maps the covariance matrix $Cov$ to a real number. Such mapping is given by the $A-$ criterion. $A-$ criterion is defined as function of the information matrix, corresponding to the associated statistical model. it suppresses the total variance of the parameters without consideration for the correlations between these estimates. This criterion is particularly advantageous if all parameters are equally important for the optimization. $A-$ criterion is proportional to the average semiaxis length of the confidence ellipsoid (cf. Figure 3). Compared to other criteria, $A-$ criterion selects the most probable value of the numerical range. (for an overview of various criteria, see [36]).

$A-$ criterion is defined as following

$$\Phi = \frac{1}{n}\text{trace}(C).$$

(17)

Now if we recall the Equation (10)

$$\tilde{N} = \mathcal{g}(S, I_{true}) = \arg\min_{\tilde{N}} \underbrace{\left\|I - S\tilde{N}\right\|^2}_{F(\hat{I},\tilde{N},S)} = \underbrace{\left\|\hat{I} - m(\tilde{N}, S)\right\|^2}_{F}.$$

(18)

The solution of Equation (18) is discussed in [5] and given by

$$(J^T J)^{-1} J\ F = \mathcal{g}(S, I_{true}) = \tilde{N},$$

(19)

where $J$ is Jacobian matrix defined as $J = \frac{\partial F}{\partial \tilde{N}}$.

With $K = \frac{\partial \mathcal{g}}{\partial \hat{I}} = (J^T J)^{-1} J^T \cdot \Sigma_I^{-1}$ we can rewrite Equation (15) as following

$$C = (J^T J)^{-1} J^T\ \Sigma_I^2\ J(J^T J)^{-1}$$

$$C = (J^T J)^{-1}.$$

$\Phi$ depends on $S$, it follows

$$\min_{S \in \mathbb{R}^{m \times n}} \Phi(S) = \text{trace}((S^T S)^{-1}).$$

(20)

Now we recall the Equation (9) with $N = \widetilde{N}/|\widetilde{N}|$. The Taylor series expansion of this term is

$$N(\widetilde{N}) = N(\widehat{N}) + \underbrace{\frac{\partial N}{\partial \widetilde{N}}(\widehat{N})}_{B}(\widetilde{N} - \widehat{N}) + \underbrace{O\left(\|\widetilde{N} - \widehat{N}\|^2\right)}_{neglect\ in\ the\ following} \quad . \tag{21}$$

$$B = \frac{\partial}{\partial \widetilde{N}}\left(\frac{\widetilde{N}}{\|\widetilde{N}\|}\right) = \begin{pmatrix} \frac{\partial}{\partial \widetilde{N}_1}\left(\frac{\widetilde{N}_1}{\|\widetilde{N}\|}\right) & \cdots & \frac{\partial}{\partial \widetilde{N}_k}\left(\frac{\widetilde{N}_1}{\|\widetilde{N}\|}\right) \\ \vdots & \ddots & \vdots \\ \frac{\partial}{\partial \widetilde{N}_1}\left(\frac{\widetilde{N}_l}{\|\widetilde{N}\|}\right) & \cdots & \frac{\partial}{\partial \widetilde{N}_k}\left(\frac{\widetilde{N}_l}{\|\widetilde{N}\|}\right) \end{pmatrix} . \tag{22}$$

$B$ is a matrix including the inaccurate normal vector $\widetilde{N}$; $\widetilde{N}$ is computed using the basic PS Equation (4). With $B$ we can make the optimization problem depending on the shape of the object we strive to measure. If we integrate $B$ in Equation (20), we obtain

$$\min_{S \in \mathbb{R}^{m \times n}} \Phi(S) = \text{trace}(B \cdot (S^T S)^{-1} \cdot B^T) . \tag{23}$$

Equation (23) can be used to calculate the optimal positions of light sources, which in turn is utilized to estimate the surface normal vector applying PS. A numerical solution for this equation is shown in Appendix A.

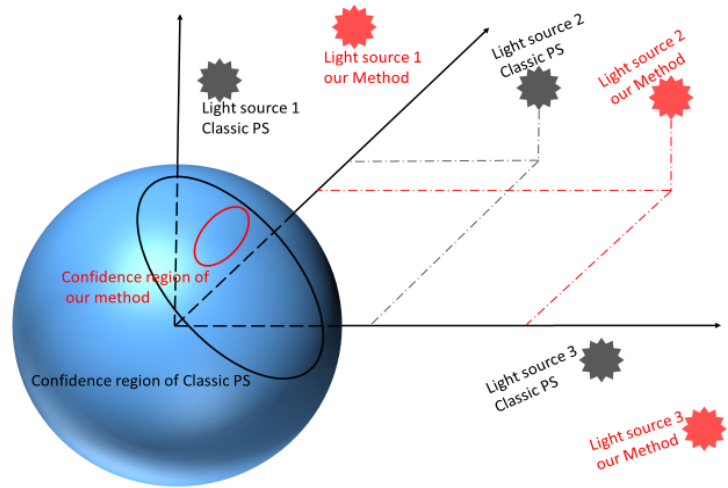

Figure 4 *Schematic representation of our approach using the unit sphere. The Optimal position and arrangement of the light sources is calculated a using confidence region. While the confidence region for classic PS is large(black), by applying our method, the confidence region is reduced (red). This allows us to calculate exact positions of the light sources so that the surface normal estimated with the highest achievable accuracy.*

## 4. Experiments

We characterize the performance of our method using both synthetic and real data-sets. The focus of this article is to determine the optimal configuration of the light sources for given PS approach and using this information to compute the normal vector. For qualitative evaluation, we took three images only, as is the case with classic PS. We compare the results of our method against classic PS and the method in [22]. Only for the synthetic data-sets is ground truth (GT) available for a quantitative analysis of our approach. Depending on proposition 3.1 in [37], if the light directions and intensities are known in photometric stereo, then the reconstructed normal map for both models (perspective projection and orthogonal projection) is the same.

### 4.1. Synthetical Images

To validate our analysis, we ran quantitative experiments on images synthetically generated with Blender software [38] under the perspective model with two given objects: Teddy and Vase (the objects are available on BlendSwap [39] and [40]). The objects are illuminated with a point light source in three different positions. For classic PS, the positions of the light source are randomly selected; for our method, the positions are calculated using Equation (23). In Figure 5 we show the objects and corresponding normal map.

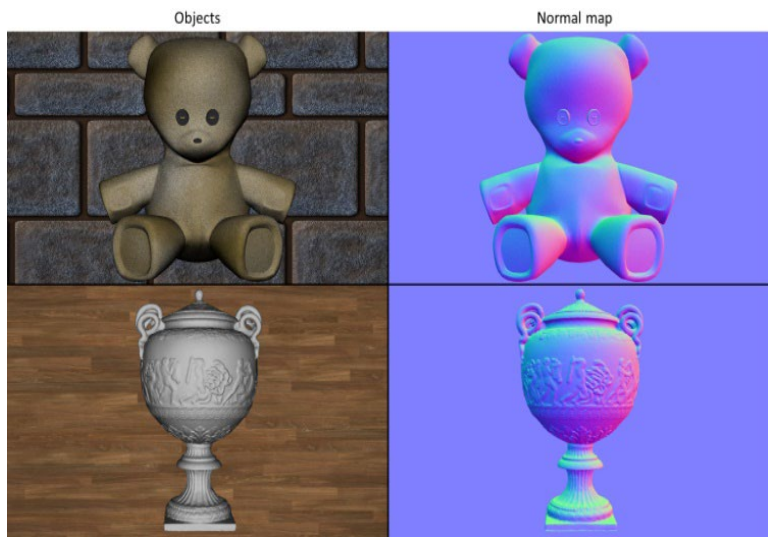

*Figure 5 Synthetic objects and corresponding normal map.*

In figure 6, we show the results of our method compared with photometric stereo. At the first glance, the results of the both methods seem to be comparable. However, a direct comparison between zoomed area on the normal map of classic PS and our method shows that using our method reveals better and sharper surface structures; even in an area where the angular error is not too large by using classic PS (cf. Figure 7).

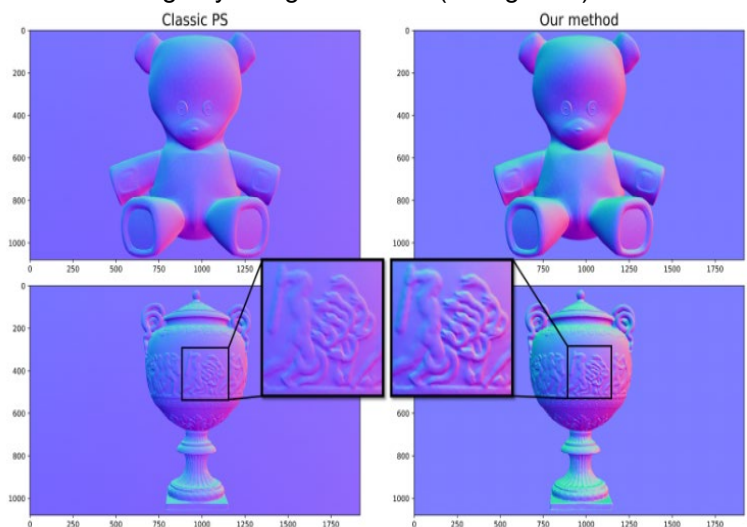

*Figure 6 Normal map of Teddy and Vase; (left) normal map computed using Classic PS approach, (right) normal map computed using our method. Even in an area where the angular error by using classic PS is not to large (cf. figure 7), using our method will reveal better and sharper surface structures.*

To demonstrate the effectiveness of our method with synthetic datasets: for each pixel, we compared the calculated angles between the direction of normal vector of our method and classic PS method with the ground-truth. To receive an impression regarding the performance of our method, we show in Figure 7 the angular error in range of $10°$ between our method and GT and between classic PS and GT. As we see, the angular error of our method is significantly lower than classic PS. Nearly everywhere our method obtains a more favorable result than classic PS. The area with the angular error bigger than $\geq 10°$ in our method occurs only at the edges. This is explained by the fact that the edges cannot be captured correctly by the camera or in these areas no intensity changes can be observed.

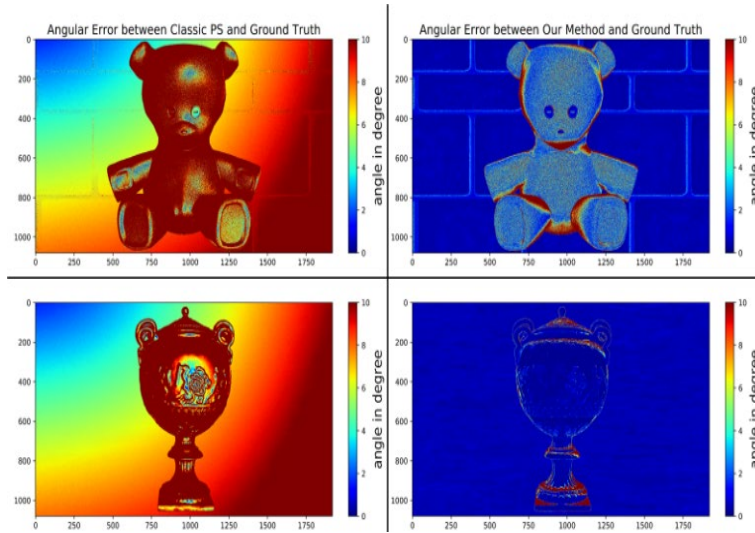

*Figure 7 Comparison between normal maps; Angular Error in degree between ground-truth and classic PS (left) and ground-truth and our method (right).*

In Figure 8 we demonstrate the distribution of the angular error of the normal vector using our method, classic PS and the method of Drbohlav and Chantler [22] for illumination configuration in PS. By looking at the results, the error distribution by using our method is close to zero, whereas the classic PS is between $5°\ to\ 10°$ and Drbohlav´s and Chantler´s [22] method between $3°\ to\ 5°$. The distribution of the angle error using our method has a mean of $2.010°$ and a median of $0.593°$, using classic PS the mean is $9.712°$ and the median $8.733°$ and using the method in [22] the mean is $6.078°$ and the median $3.313°$. In this way we also demonstrate the robustness and accuracy of our method.

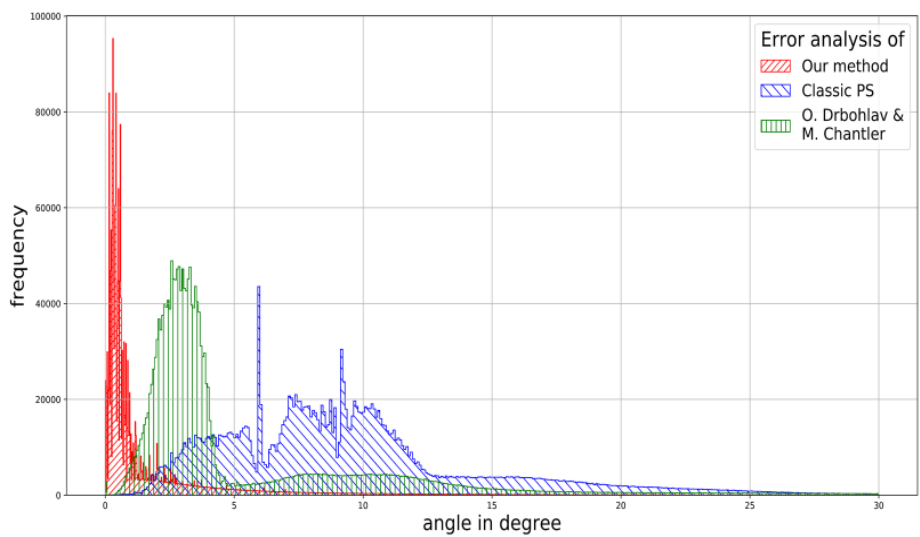

*Figure 8 Error analysis of the normal vector using classic PS (blue), the method presented in [22] (green) and our method (red).*

To evaluate the performance of our method and for the avoidance of doubt by selecting bad positions of the light source for our experiment, we generated $1e^6$ randomly positions of the light sources distributed with a discrete uniform distribution. We then compared the measure of the size of the confidence region $\Phi(S)$, as defined by Equation (23), of randomly generated positions and of our method Even if the selection of the light sources obeys a heuristic (e.g, maximize the distance between the light sources) our approach still provides a better configuration of the light to calculate surface normal cf. Figure 9.

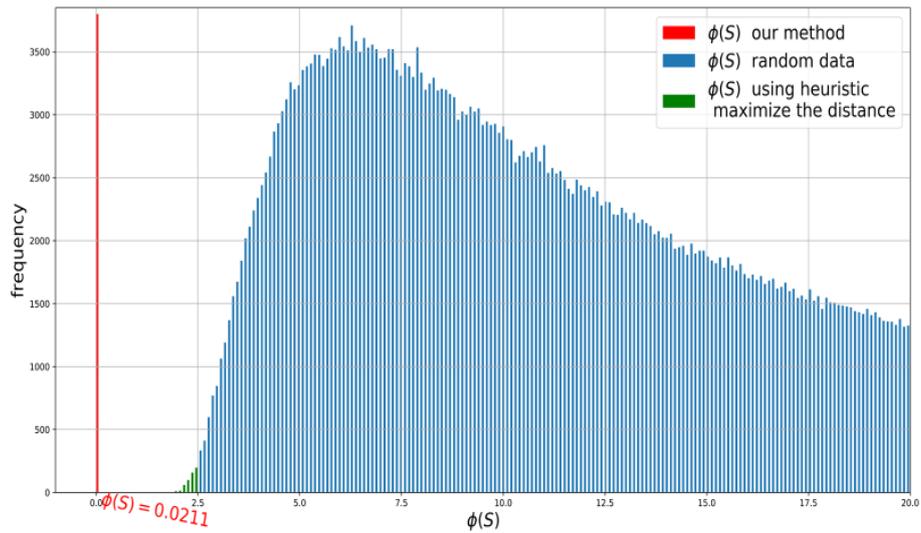

*Figure 9* Distribution of $\Phi(S)$. The Measure of the size of confidence region $\Phi(S)$ of $1e^6$ randomly generated data (blue), random data using heuristic (maximize the distance between the light sources) and our method (red). In comparison to random data and random data using heuristic (maximize the distance between the light sources), with our method $\Phi(S)$ reaches a value of 0.0211 while the best $\Phi(S)$ of random data using heuristic (maximize the distance between the light sources) is 1.8657.

### 4.2. Real Images

To evaluate the efficiency of our method in a more realistic setting, in this section we demonstrate experimental results for three real objects: a flower pot, a kettlebell weight and an engine part. Three-color images of each object captured under different illumination directions and used PS technique to calculate the corresponding normal map. The light source is mounted on a robot arm, allowing accurate movement of the light source. An experimental setup can be seen in Figure10.

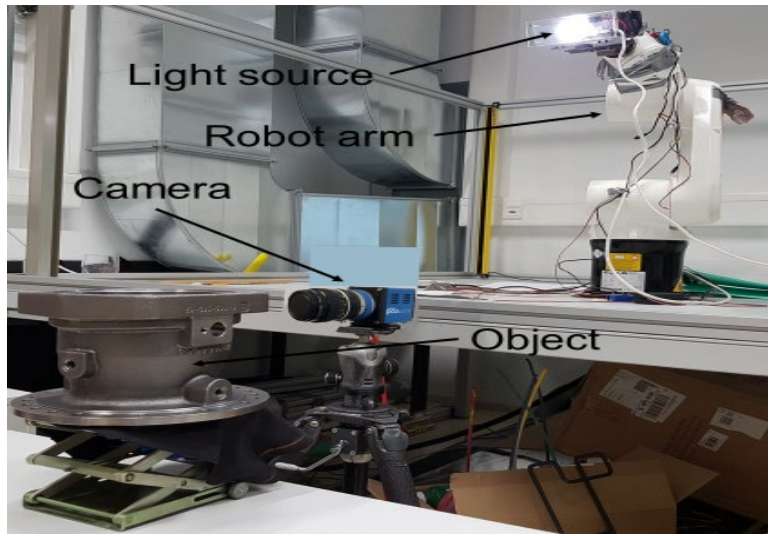

*Figure 10 Overview of our experimental setup using robot arm to move light source.*

Since no ground-truth values are available for real data set, we only show the visual improvements. The experimental results of our method are consistent with those of a classic PS approach. The qualitative comparison of the two methods is provided in Figure11. Inspection of the results reveals that the calculated normal map, using our method, is more detailed than the classic photometric stereo and contains sharper surface structures (cf. selected image areas on Figure 11). This feature enables the use of our method in automated visual inspection and all other tasks where precision in the approximation of the object surface and flexibility play a significant role (i.e. in industry 4.0).
By comparing the results, it is striking that our method in contrast to classic PS, is more stable to noise (especially camera noise) which in turn underlines the stability of our method.

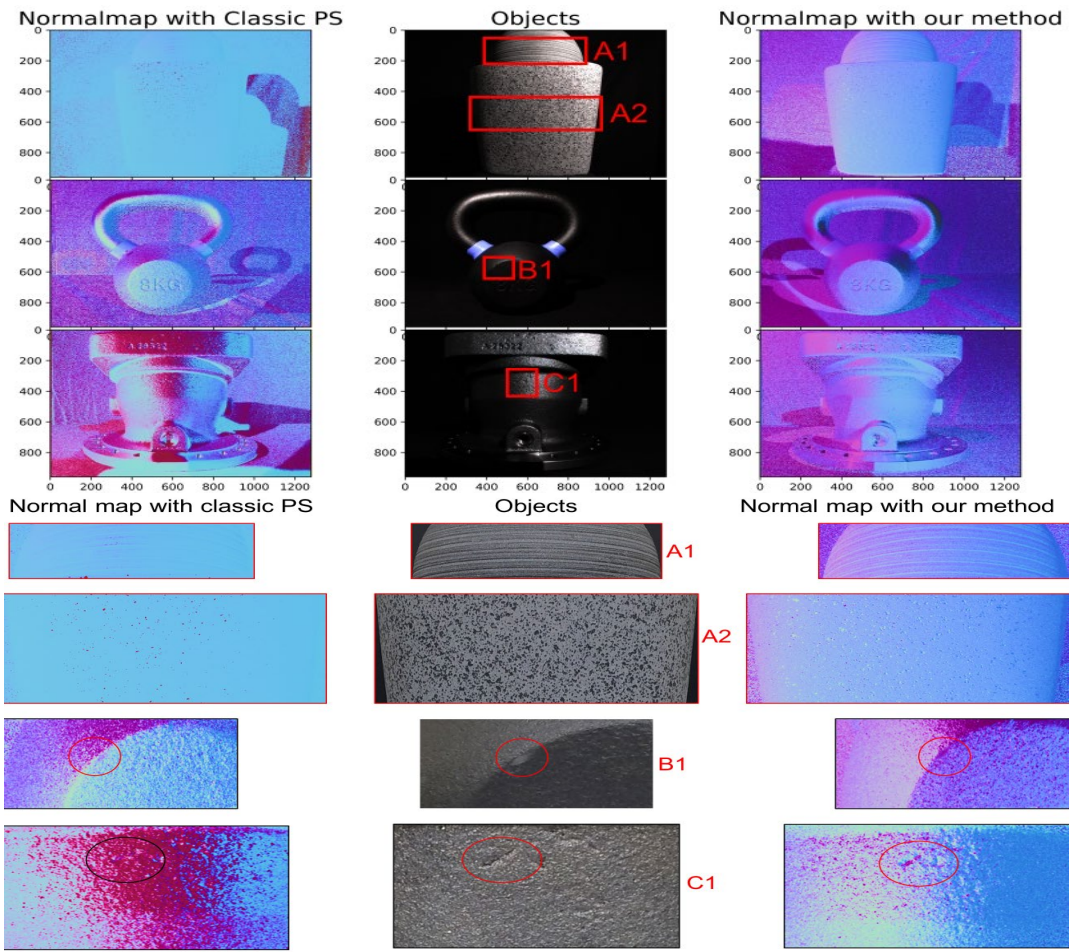

*Figure 11 Normal map of three real objects. Comparison between results of classic PS (left) and our method (right). To visualize the difference, the selected image areas on the upper image are zoomed in the lower image. The selected image areas underline the precision of our method in calculating the normal map compared with classic PS.*

5. Conclusions

In many applications, the surface normal is an important basis for robust 3D features, for example in surface orientation and curvature. Surface orientation at a point on the object surface can be computed from image intensities obtained under a fixed imaging geometry but with different light conditions. In this paper, we have analyzed the effect of positioning the light sources on the quality of the normal vector for PS and used this knowledge to introduce a novel approach that combines methods of calibrated photometric stereo with model-based parameter estimation and optimum experimental design for estimating surface normal. The focus of this study was to find out the optimal configuration of the light sources to estimate the normal map using the methods of parameter estimation, optimum experimental design and photometric stereo. The input to our approach is a set of images taken under variable point light sources and a fixed camera position. The results demonstrate that the method can work well on synthetic scenes with Lambertian BRDF as well as on real-world objects. The accuracy of our method for estimating the normal vector is better than both classic PS and the state of art methods (e.g., [22]), we have proved and demonstrated this with experiments. When applying our method, the mean of the angle error distribution is $2.010°$, while using the method [22] is $6.078°$ and using the classic PS is $9.712°$. PS methods are independent of the geometry of the object surface; by applying our approach for more accuracy by estimation of the normal vector, better surface reconstruction on the one hand and on the other better estimation of Bidirectional reflectance distribution function (BRDF) can be archived. We have illustrated that, by applying our approaches to compute the optimal configuration of the light sources, the results of photometric stereo are significantly improved. Our method therefore can be expanded to all techniques or approaches, in which lighting is important to achieve better results. It can be both geometrical methods as well as photometric methods, with the exceptions of diffuse light, in the present of cast shadow and inter reflections. In such cases, the described approach is inoperative. The future research will focus on broadening this approach so that it can deals with such effects.

# 6. Appendices
## 6.1. Appendix A *(a numerical solution to Equation 23)*

In this part we present a possible numerical solution to Equation 23. Let us recall this equation

$$\min_{S\in\mathbb{R}^{m\times n}} \Phi(S) = \text{trace}(B \cdot (S^T S)^{-1} \cdot B^T),$$

according to cyclic invariance of the trace

$$\cong \min_{S\in\mathbb{R}^{m\times n}} \text{trace}((B^T B) \cdot (S^T S)^{-1}).$$

By means of automatic differentiation showing in [41] we can solve this equation in the following steps. According to appendix B for idempotent symmetric matrix, it follows that $B^T \cdot B = B$.

Forward mode:

$$\begin{aligned}
v_1 &= (S^T S) \\
v_2 &= \text{inverse}(v_1) \\
v_3 &= B \qquad \text{(according to appendix B)} \\
v_4 &= v_2 \\
v_5 &= v_3 \cdot v_4 \\
v_6 &= sum(v_5) \\
y &= v_6.
\end{aligned}$$

Reverse mode:

$$\begin{aligned}
\bar{v}_6 &= 1 \\
\bar{v}_5 &= 1 \\
\bar{v}_4 &\mathrel{+}= \bar{v}_5 \cdot v_3 \\
\bar{v}_3 &\mathrel{+}= \bar{v}_5 \cdot v_4 \\
\bar{v}_2 &\mathrel{+}= \bar{v}_4 \\
\bar{v}_1 &\mathrel{+}= -(v_2^T \bar{v}_2 v_2^T) \\
\bar{s} &= S_{initial} \cdot (\bar{v}_1^T + \bar{v}_1),
\end{aligned}$$

where $\bar{s} = \nabla_S \Phi(S) \in \mathbb{R}^{3\times 3}$, this gradient is required for a gradient descent optimization method [42].

## 6.2. Appendix B *(to show that $B$ is symmetric and idempotent)*

**Lemma**: Let $\widetilde{N} \in \mathbb{R}^n$ then it follows that $B = \frac{\partial}{\partial \widetilde{N}}\left(\frac{\widetilde{N}}{\|\widetilde{N}\|}\right)_{k,l}$ is symmetric and idempotent.

To show:    1. $B^T = B$,
              2. $B \cdot B = B^2 = B$.

**Proof** First, we show that the matrix $B$ a symmetric matrix. $B$ is symmetric if and only if

$$B_{k,l} = B_{l,k} \qquad \forall_{k,l} \in \mathbb{R}$$

for $k = l$ it is a diagonal matrix for $k \neq l$

$$B_{k,l} := \frac{\partial}{\partial \widetilde{N}_k}\left(\frac{\widetilde{N}_l}{\|\widetilde{N}\|}\right) = \frac{\partial}{\partial \widetilde{N}_k} \frac{\widetilde{N}_l}{\sqrt{\widetilde{N}_1^2 + \cdots + \widetilde{N}_k^2}} = \frac{-\widetilde{N}_l \cdot \widetilde{N}_k}{\sqrt{\left(\widetilde{N}_1^2 + \cdots + \widetilde{N}_k^2\right)^3}} = \frac{\partial}{\partial \widetilde{N}_l}\left(\frac{\widetilde{N}_k}{\|\widetilde{N}\|}\right) = \frac{\partial}{\partial \widetilde{N}}\left(\frac{\widetilde{N}}{\|\widetilde{N}\|}\right)_{l,k} = B_{l,k}.$$

Now to prove $B \cdot B = B^2 = B$. According to theorem 10.2 in [43] the following holds $B$ is idempotent if and only if $\text{trace}(B) = \text{rank}(B)$. It remains to show that for our $B$ it holds $\text{trace}(B) = \text{rank}(B)$.
If $B = 0$ is trivial; then $\text{trace}(B) = 0 = \text{rank}(B)$.
Now let us observe the case where $B$ is nonnull. Let $n$ denote the order of $B_{n\times n}$, and let $r = \text{rank}(B)$. Then according to Rank Factorization chapter 4 of [43], there exists matrices $L_{n\times r}$ and $R_{r\times n}$ in a way that $B_{n\times n} = L_{n\times r} R_{r\times n}$. Where $L$ is of full column rank and $R$ is of full row rank, then $L$ has left inverse $L^\star$ and $R$ has right inverse $R^\star$.

Now, since $B^2 = B$, we have

$$LR\, LR = LR$$

$$\underbrace{L^\star L}_{Id}\, R\ LR = \underbrace{L^\star L}_{Id} \quad \text{| multiply with } L^\star \text{ from left}$$

$$RL\, \underbrace{RR^\star}_{Id} = \underbrace{RR^\star}_{Id} \text{ | multiply with } R^\star \text{ from right}$$

$$RL = Id_{r\times r},$$

therefore

$$\text{trace}(B) = \text{trace}(LR) = \text{trace}(RL) = \text{trace}(Id_{r\times r}) = r = \text{rank}(B) \quad \blacksquare$$

where $Id$ is identity matrix.

## Acknowledgment

The authors would like to express their thanks to the company Pulse Getriebe for providing the test object (engine part), in addition, we thank Benjamin Reh at Robotic Teaching Lab (ORB-Uni Heidelberg) for providing the robot system.